\def\year{2020}\relax
\documentclass[letterpaper]{article} %
\usepackage{aaai20}  %
\usepackage{times}  %
\usepackage{helvet} %
\usepackage{courier}  %
\usepackage[hyphens]{url}  %
\usepackage{graphicx} %
\usepackage{graphicx}  %
\usepackage{amsmath}
\usepackage{amssymb}
\usepackage{multirow}
\usepackage{hyperref}
\title{Context-Transformer: Tackling Object Confusion for Few-Shot Detection}
\author{Ze Yang, \textsuperscript{\rm 1} \thanks{Ze Yang and Yali Wang contribute equally}
	Yali Wang,\textsuperscript{\rm 1} \footnotemark[1]
  Xianyu Chen,\textsuperscript{\rm 1}
  Jianzhuang Liu,\textsuperscript{\rm 2}
	Yu Qiao \textsuperscript{\rm 1 3}\thanks{Corresponding author.}   \\
	\textsuperscript{\rm 1}ShenZhen Key Lab of Computer Vision and Pattern Recognition, SIAT-SenseTime Joint Lab,\\ Shenzhen Institutes of Advanced Technology, Chinese Academy of Sciences\\ %
	\textsuperscript{\rm 2}Huawei Noah's Ark Lab\\ %
	\textsuperscript{\rm 3}SIAT Branch, Shenzhen Institute of Artificial Intelligence and Robotics for Society\\ %
	\{ze.yang, yali.wang, yu.qiao\}@siat.ac.cn,
   xianyuchen1992@outlook.com,
   liu.jianzhuang@huawei.com
}

\begin{document}

\maketitle

\begin{abstract}
Few-shot object detection is a challenging but realistic scenario,
where only a few annotated training images are available for training detectors.
A popular approach to handle this problem is transfer learning,
i.e.,
fine-tuning a detector pretrained on a source-domain benchmark.
However,
such transferred detector often fails to recognize new objects in the target domain,
due to low data diversity of training samples.
To tackle this problem,
we propose a novel Context-Transformer within a concise deep transfer framework. %
Specifically,
Context-Transformer can effectively leverage source-domain object knowledge as guidance,
and automatically exploit contexts from only a few training images in the target domain.
Subsequently,
it can adaptively integrate these relational clues to enhance the discriminative power of detector,
in order to reduce object confusion in few-shot scenarios.
Moreover,
Context-Transformer is flexibly embedded in the popular SSD-style detectors,
which makes it a plug-and-play module for end-to-end few-shot learning.
Finally,
we evaluate Context-Transformer on the challenging settings of few-shot detection and incremental few-shot detection.
The experimental results show that,
our framework outperforms the recent state-of-the-art approaches.
The codes are available at \url{https://github.com/Ze-Yang/Context-Transformer}.
\end{abstract}

\begin{figure*}[t]
\centering
\includegraphics[width=1.6\columnwidth]{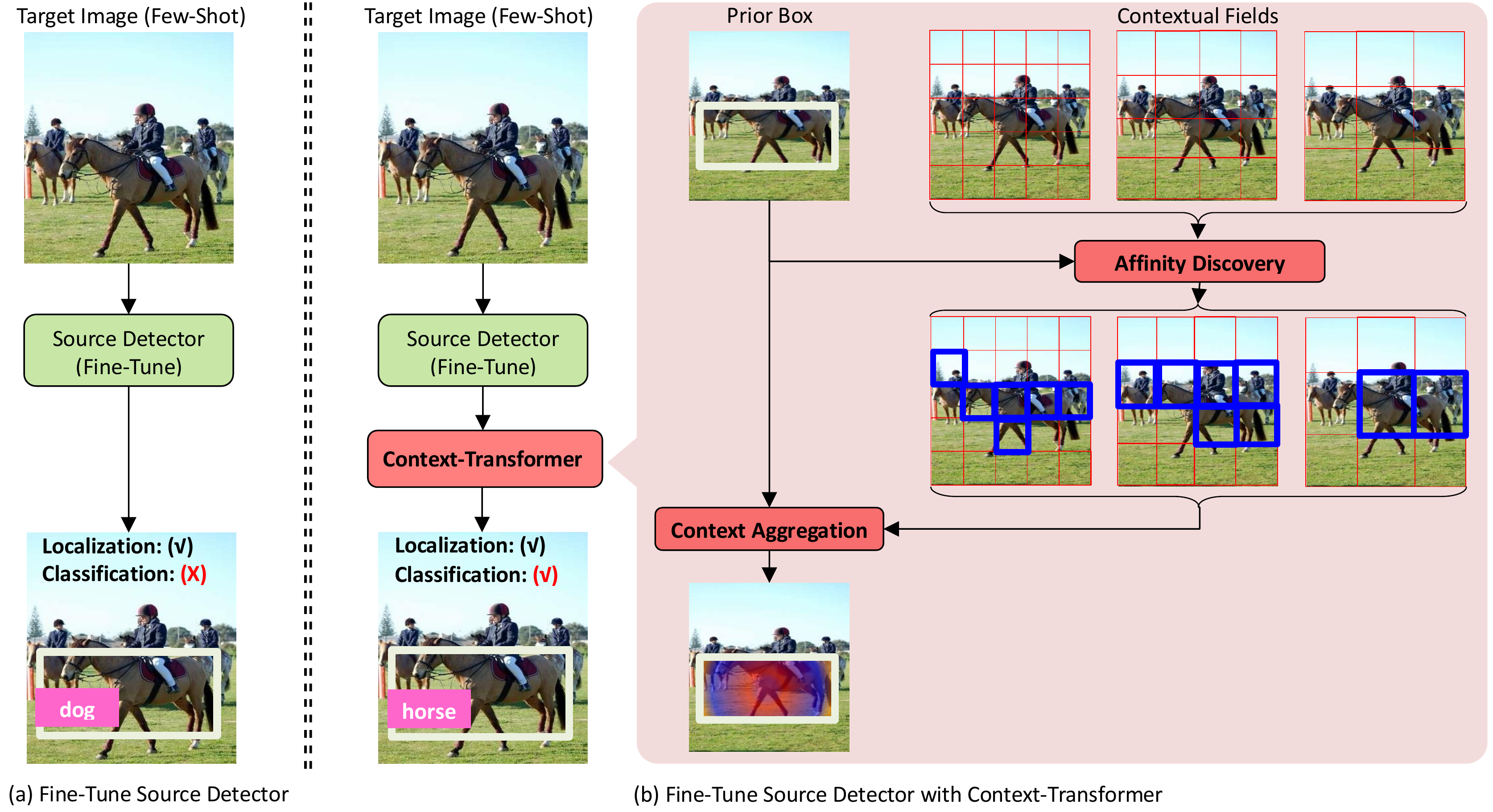}
\caption{Our Motivation.
Fine-tuning a pretrained detector is a popular approach for few-shot object detection.
However,
such transferred detector often suffers from object confusion in the new target domain,
e.g.,
a \textit{horse} is misclassified as a \textit{dog},
due to annotation scarcity.
Alternatively,
humans can effectively correct such few-shot confusions by further exploiting discriminative context clues from only a few images on hand.
Inspired by this observation,
we introduce a novel Context-Transformer to tackle object confusion for few-shot object detection.
More explanations can be found in Section \ref{Introduction} and \ref{MetaTransformer}.}
\label{Mot}
\end{figure*}

\section{Introduction}
\label{Introduction}

Object detection has been mainly promoted by deep learning frameworks \cite{ren2015faster,he2017mask,redmon2016you,liu2016ssd}.
However,
the impressive performance of these detectors heavily relies on large-scale benchmarks with bounding box annotations,
which are time-consuming or infeasible to obtain in practice.
As a result,
we often face a real-world scenario,
i.e.,
few-shot object detection,
where there are only a few annotated training images.
In this case,
deep learning will deteriorate due to severe overfitting.

A popular strategy is transfer learning,
i.e.,
one can train an object detector with a large-scale benchmark in the source domain,
and then fine-tune it with a few samples in the target domain.
By doing so,
we observe an interesting and important phenomenon.
For few-shot object detection,
a transferred detector often performs well on localization while encounters difficulty in classification,
e.g.,
a \textit{horse} is well-localized but misclassified as a \textit{dog} in Fig. \ref{Mot}.

The main reason is that,
an object detector uses bounding box regressor (BBOX) for localization while object+background classifier (OBJ+BG) for classification.
BBOX is often category-irrelevant.
Hence,
we can use source-domain BBOX as a reliable initialization of target-domain BBOX.
In this case,
the detector can effectively localize new objects after fine-tuning with a few training samples in the target domain.
On the contrary,
OBJ+BG is category-specific.
In other words,
it has to be randomly initialized for new categories in the target domain.
However,
only a few training images are available in this domain.
Such low data diversity significantly enlarges the training difficulty of classifier,
which leads to the key problem above,
i.e.,
object confusion caused by annotation scarcity.

To address this problem,
we propose a novel Context-Transformer.
It can automatically exploit contexts from only a few images on hand,
and attentively integrate such distinct clues to generalize detection.
Our design is inspired by \cite{oliva2007role} that, at an early age with little object knowledge, humans can build the contextual associations for visual recognition. In other words, under little supervision scenarios, we will try to explore distinct clues in the surroundings (which we refer to as contextual fields in this paper), to clarify object confusion.
For example,
a few images may be discriminative enough to distinguish \textit{horse} from \textit{dog},
when we find that these images contain important contents such as a person sits on this animal, the scene is about wild grassland, etc.

To mimic this capacity,
we design Context-Transformer in a concise transfer framework.
Specifically,
it consists of two simple but effective submodules,
i.e.,
affinity discovery and context aggregation.
For a target-domain image,
affinity discovery first constructs a set of contextual fields,
according to default prior boxes (also called as anchor boxes) in the detector.
Then,
it adaptively exploits relations between prior boxes and contextual fields.
Finally,
context aggregation leverages such relations as guidance,
and integrates key contexts attentively into each prior box.
As a result,
Context-Transformer can generate a context-aware representation for each prior box,
which allows detector to distinguish few-shot confusion with discriminative context clues.
\textbf{To our best knowledge,
Context-Transformer is the first work to investigate context for few-shot object detection}.
Since it does not require excessive contextual assumptions on aspect ratios, locations and spatial scales,
Context-Transformer can flexibly capture diversified and discriminative contexts to distinguish object confusion.
More importantly,
it leverages elaborative transfer insights for few-shot detection.
With guidance of source-domain knowledge,
Context-Transformer can effectively reduce learning difficulty when exploiting contexts from few annotated images in the target domain.
Additionally,
we embed Context-Transformer into the popular SSD-style detectors.
Such plug-and-play property makes it practical for few-shot detection.
Finally,
we conduct extensive experiments on different few-shot settings,
where our framework outperforms the recent state-of-the-art approaches.

\section{Related Works}

Over the past years,
we have witnessed the fast development of deep learning in object detection.
In general,
the deep detection frameworks are mainly categorized into two types,
i.e.,
one-stage detectors (e.g., YOLO or SSD styles \cite{redmon2016you,liu2016ssd})
and
two-stage detectors (e.g., R-CNN styles \cite{girshick2014rich,girshick2015fast,ren2015faster,he2017mask}).
Even though both types have achieved great successes in object detection,
they heavily depend on large-scale benchmarks with bounding boxes annotations.
Collecting such fully-annotated datasets is often difficult or labor-intensive for real-life applications.

\textbf{Few-Shot Object Detection}.
To alleviate this problem,
weakly~\cite{bilen2016weakly,lai2017saliency,tang2017multiple} or semi~\cite{hoffman2014lsda,tang2016large} supervised detectors have been proposed.
However,
only object labels are available in the weakly-supervised setting,
which restricts the detection performance.
The semi-supervised detectors often assume that,
there is a moderate amount of object box annotations,
which can be still challenging to obtain in practice.
Subsequently,
few-shot data assumption has been proposed in \cite{dong2018few}.
However,
it relies on multi-model fusion with a complex training procedure,
which may reduce the efficiency of model deployment for a new few-shot detection task.
Recently,
a feature reweighting approach has been introduced in a transfer learning framework \cite{kang2019few}.
Even though its simplicity is attractive,
this approach requires object masks as extra inputs to train a meta-model of feature reweighting.
More importantly,
most approaches may ignore object confusion caused by low data diversity.
Alternatively,
we propose a novel Context-Transformer to address this problem.

\textbf{Object Detection with Contexts}.
Modeling context has been a long-term challenge for object detection \cite{bell2016inside,chen2017spatial,kantorov2016contextlocnet,mottaghi2014role}.
The main reason is that,
objects may have various locations, scales, aspect ratios, and classes.
It is often difficult to model such complex instance-level relations by manual design.
Recently,
several works have been proposed to alleviate this difficulty,
by automatically building up object relations with non-local attention \cite{wang2018non,hu2018relation}.
However,
these approaches would lead to unsatisfactory performance in few-shot detection,
without elaborative transfer insights.
Alternatively,
our Context-Transformer is built upon a concise transfer framework,
which can leverage source-domain object knowledge as guidance,
and effectively exploit target-domain context for few-shot generalization.

\textbf{Few-Shot Learning}.
Unlike deep learning models,
humans can learn new concepts with little supervision \cite{lake2015human}.
For this reason,
few-shot learning has been investigated by
Bayesian program learning~\cite{lake2015human},
memory machines \cite{graves2014neural,santoro2016meta},
meta learning \cite{finn2017model,yoon2018bayesian},
metric learning \cite{qi2018low,snell2017prototypical,vinyals2016matching},
etc.
However,
these approaches are designed for the standard classification task.
Even though recently few-shot semantic segmentation is investigated by \cite{zhang2019canet,zhang2019pyramid}, such task does not consider intra-class instance separation.
Hence,
they may lack the adaptation capacity for few-shot object detection.

\section{Source Detection Transfer}
\label{Source Detection Transfer}

To begin with,
we formulate few-shot object detection in a practical transfer learning setting.
\textbf{First},
we assume that,
we can access to a published detection benchmark with $C_{s}$ object categories.
It is used as large-scale dataset for model pretraining in the source domain.
\textbf{Second},
we aim at addressing few-shot detection in the target domain.
Specifically,
this task consists of $C_{t}$ object categories.
For each category,
there are only $N$ fully-annotated training images,
e.g.,
$N$=5 for 5-shot case.
\textbf{Finally},
we consider a challenging transfer scenario,
i.e.,
object categories are non-overlapped between source and target domains,
for evaluating whether our framework can generalize well on new object categories.

\textbf{Detection Backbone}.
In this work,
we choose the SSD-style detector \cite{liu2016ssd,liu2018receptive} as backbone.
One reason is that,
multi-scale spatial receptive fields in this architecture provide rich contexts.
Additionally,
its concise detection design promotes flexibility of our transfer framework in practice.
In particular,
the SSD-style detector is a one-stage detection framework,
which consists of detection heads on $K$ spatial scales.
For each spatial scale,
the detection heads contain bounding box regressor (BBOX) and object+background classifier (OBJ+BG).

\textbf{Source Detection Transfer}.
To generalize few-shot learning in the target domain,
we first pretrain the SSD-style detector with large-scale benchmark in the source domain.
In the following,
we explain how to transfer source-domain detection heads (i.e., BBOX and OBJ+BG),
so that one can leverage prior knowledge as much as possible to reduce overfitting for few-shot object detection.

\textbf{(1) Source BBOX: Fine-Tuning}.
BBOX is used for localization.
As it is shared among different categories,
source-domain BBOX can be reused in the target domain.
Furthermore,
source-domain BBOX is pretrained with rich annotations in the large-scale dataset.
Hence,
fine-tuning this BBOX is often reliable to localize new objects,
even though we only have a few training images in the target domain.

\textbf{(2) Source BG: Fine-Tuning}.
OBJ+BG is used for classification.
In this work,
we factorize OBJ+BG separately into OBJ and BG classifiers.
The reason is that,
BG is a binary classifier (object or background),
i.e.,
it is shared among different object categories.
In this case,
the pretrained BG can be reused in the target domain by fine-tuning.

\textbf{(3) Source OBJ: Preserving}.
The last but the most challenging head is OBJ,
i.e.,
multi-object classifier.
Note that,
object categories in the target domain are non-overlapped with those in the source domain.
Traditionally,
one should unload source-domain OBJ and add a new target-domain OBJ.
However,
adding new OBJ directly on top of high-dimensional feature would introduce a large number of randomly-initialized parameters,
especially for multi-scale design in SSD-style frameworks.
As a result,
it is often hard to train such new OBJ from scratch,
when we have only a few annotated images in the target domain.
Alternatively,
we propose to preserve source-domain OBJ and add a new target-domain OBJ on top of it.
The main reason is that,
the dimensionality of prediction score in source-domain OBJ is often much smaller than the number of feature channels in convolutional layers.
When adding a new target-domain OBJ on top of source-domain OBJ,
we will introduce fewer extra parameters and therefore alleviate overfitting.

\textbf{Context-Transformer Between Source and Target OBJs}.
To some degree,
preserving source-domain OBJ can reduce the training difficulty of target-domain OBJ.
However,
simple source detection transfer is not enough to address the underlying problem of few-shot object detection,
i.e.,
object confusion introduced by annotation scarcity in the target domain.
Hence,
it is still necessary to further exploit target-domain knowledge effectively from only a few annotated training images.
As mentioned in our introduction,
humans often leverage contexts as a discriminative clue to distinguish such few-shot confusion.
Motivated by this,
we embed a novel Context-Transformer between source and target OBJs.
It can automatically exploit contexts,
with the guidance of source domain object knowledge from source OBJ.
Then,
it can integrate such relational clues to enhance target OBJ for few-shot detection.

\begin{figure*}[t]
\centering
\includegraphics[width=1.7\columnwidth]{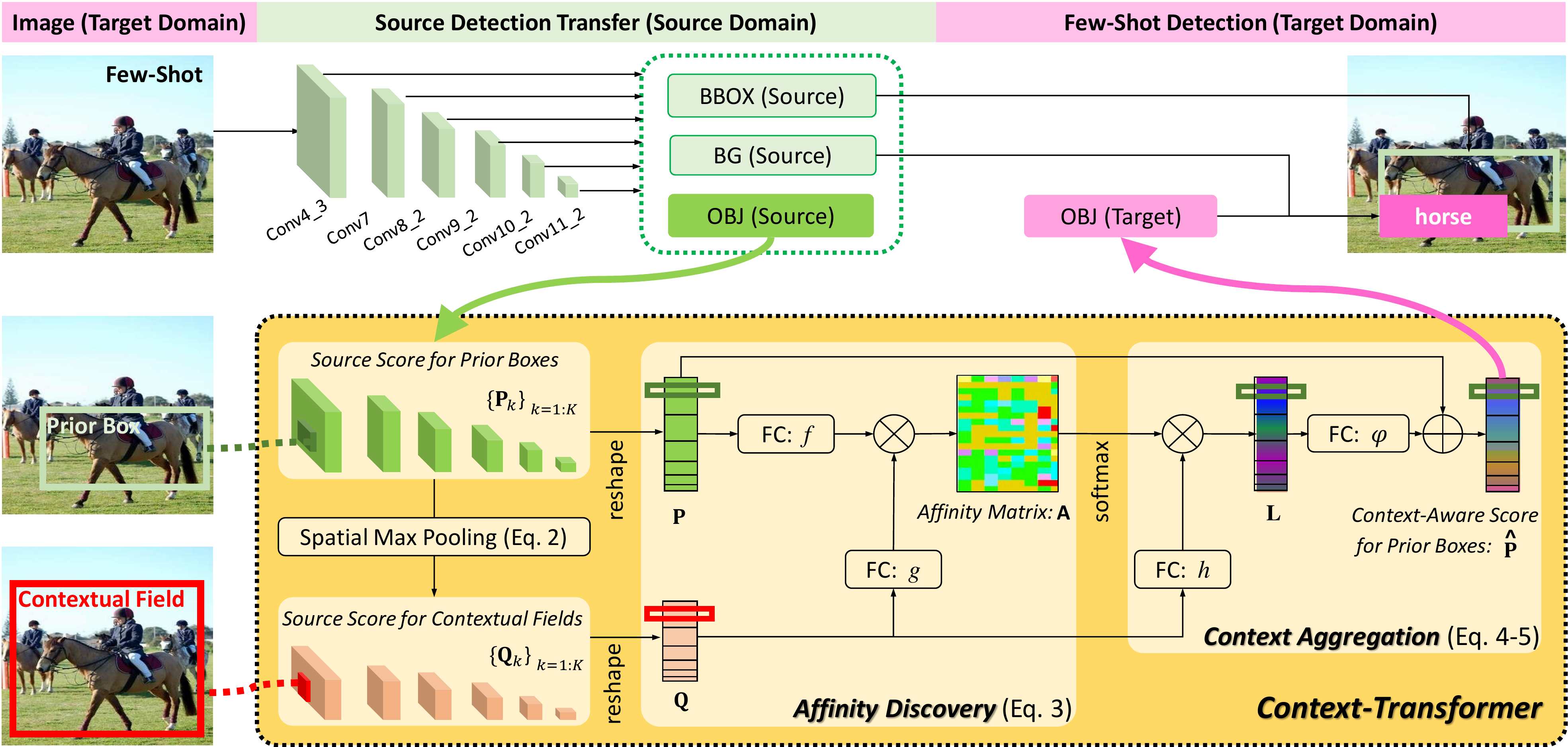}
\caption{Few-Shot Detection with Context-Transformer.
It is a plug-and-play module between source and target OBJs,
based on SSD-style detectors.
It consists of affinity discovery and context aggregation,
which can effectively reduce object confusion in the few-shot target domain,
by exploiting contexts in a concise transfer framework.
More details can be found in Section \ref{MetaTransformer}.}
\label{AllFrame}
\end{figure*}

\section{Context-Transformer}
\label{MetaTransformer}

In this section,
we introduce Context-Transformer for few-shot object detection.
Specifically,
it is a novel plug-and-play module between source and target OBJs.
We name it as Context-Transformer,
because it consists of two submodules to transform contexts,
i.e.,
affinity discovery and context aggregation.
The whole framework is shown in Fig. \ref{AllFrame}.

\subsection{Affinity Discovery}
In SSD-style detectors \cite{liu2016ssd},
prior boxes are default anchor boxes with various aspect ratios.
Since classification is performed over the representations of these boxes,
affinity discovery first constructs a set of contextual fields for prior boxes.
Subsequently,
it exploits relations between prior boxes and contextual fields in a target-domain image,
with guidance of source-domain object knowledge.

\textbf{Source-Domain Object Knowledge of Prior Boxes}.
For a target-domain image,
we should first find reliable representations of prior boxes,
in order to perform affinity discovery under few-shot settings.
Specifically,
we feed a target-domain image into the pretrained SSD-style detector,
and extract the score tensor from source OBJ (before softmax),
\begin{equation}
\mathbf{P}_{k}\in\mathbb{R}^{H_{k}\times W_{k}\times (M_{k}\times C_{s})},~~k=1,...,K,
\label{eq:SourceScore}
\end{equation}
where
$\mathbf{P}_{k}(h,w,m,:)\in\mathbb{R}^{C_{s}}$ is a source-domain score vector,
w.r.t.,
the prior box with the $m$-th aspect ratio located at $(h,w)$ of the $k$-th spatial scale.
We would like to emphasize that,
the score of source-domain classifier often provides rich semantic knowledge about target-domain object categories \cite{tzeng2015simultaneous,yim2017gift}.
Hence,
$\{\mathbf{P}_{k}\}_{k=1}^{K}$ is a preferable representation of prior boxes for a target-domain image.
Relevant visualization can be found in our supplementary material.

\textbf{Contextual Field Construction via Pooling}.
After obtaining the representation of prior boxes,
we construct a set of contextual fields for comparison.
Ideally,
we hope that contextual fields are not constructed with excessive spatial assumptions and complicated operations,
due to the fact that we only have a few training images on hand.
A naive strategy is to use all prior boxes directly as contextual fields.
However,
there are approximately 10,000 prior boxes in the SSD-style architecture.
Comparing each prior box with all others would apparently introduce unnecessary learning difficulty for few-shot cases.
Alternatively,
humans often check sparse contextual fields,
instead of paying attention to every tiny detail in an image.
Motivated by this observation,
we propose to perform spatial pooling (e.g., max pooling) over prior boxes $\mathbf{P}_{k}$.
As a result,
we obtain the score tensor $\mathbf{Q}_{k}\in\mathbb{R}^{U_{k}\times V_{k}\times (M_{k}\times C_{s})}$ for a set of contextual fields,
\begin{equation}
\mathbf{Q}_{k}=SpatialPool(\mathbf{P}_{k}),~~k=1,...,K,
\label{eq:ContextualScore}
\end{equation}
where
$U_{k}\times V_{k}$ is the size of the $k$-th scale after pooling.

\textbf{Affinity Discovery}.
To discover affinity between prior boxes and contextual fields,
we compare them according to their source-domain scores.
For convenience,
we reshape score tensors $\mathbf{P}_{1:K}$ and $\mathbf{Q}_{1:K}$ respectively as matrices $\mathbf{P}\in\mathbb{R}^{D_{p}\times C_{s}}$ and $\mathbf{Q}\in\mathbb{R}^{D_{q}\times C_{s}}$,
where
each row of $\mathbf{P}$ (or $\mathbf{Q}$) refers to the source-domain score vector of a prior box (or a contextual field).
Moreover,
$D_{p}=\sum\nolimits_{k=1}^{K}H_{k}\times W_{k}\times M_{k}$
and
$D_{q}=\sum\nolimits_{k=1}^{K}U_{k}\times V_{k}\times M_{k}$
are respectively the total number of prior boxes and contextual fields in a target-domain image.
For simplicity,
we choose the widely-used dot-product kernel to compare $\mathbf{P}$ and $\mathbf{Q}$ in the embedding space.
As a result,
we obtain an affinity matrix $\mathbf{A}\in\mathbb{R}^{D_{p}\times D_{q}}$ between prior boxes and contextual fields,
\begin{equation}
\mathbf{A}=f(\mathbf{P})\times g(\mathbf{Q})^{\top},
\label{eq:AM}
\end{equation}
where
$\mathbf{A}(i,:)\in\mathbb{R}^{1\times D_{q}}$ indicates the importance of all contextual fields,
w.r.t.,
the $i$-th prior box.
$f(\mathbf{P})\in\mathbb{R}^{D_{p}\times C_{s}}$ and $g(\mathbf{Q})\in\mathbb{R}^{D_{q}\times C_{s}}$ are embeddings for prior boxes and contextual fields respectively,
where
$f$ (or $g$) is a fully-connected layer that is shared among prior boxes (or contextual fields).
These layers can increase learning flexibility of kernel computation.
To sum up,
affinity discovery allows a prior box to identify its important contextual fields automatically from various aspect ratios, locations and spatial scales.
Such diversified relations provide discriminative clues to reduce object confusion caused by annotation scarcity.

\subsection{Context Aggregation}
\label{Context Aggregation}

After finding affinity between prior boxes and contextual fields,
we use it as a relational attention to integrate contexts into the representation of each prior box.

\textbf{Context Aggregation}.
We first add $softmax$ on each row of $\mathbf{A}$.
In this case,
$softmax(\mathbf{A}(i,:))$ becomes a gate vector that indicates how important each contextual field is for the $i$-th prior box.
We use it to summarize all the contexts $\mathbf{Q}$ attentively,
\begin{equation}
\mathbf{L}(i,:)=softmax(\mathbf{A}(i,:))\times h(\mathbf{Q}),
\label{eq:sum1}
\end{equation}
where
$\mathbf{L}(i,:)$ is the weighted contextual vector for the $i$-th prior box ($i$=$1,...,D_{p}$).
Additionally,
$h(\mathbf{Q})\in\mathbb{R}^{D_{q}\times C_{s}}$ refers to a contextual embedding,
where $h$ is a fully-connected layer to promote learning flexibility.
Finally,
we aggregate the weighted contextual matrix $\mathbf{L}\in\mathbb{R}^{D_{p}\times C_{s}}$ into the original score matrix $\mathbf{P}$,
and obtain the context-aware score matrix of prior boxes $\widehat{\mathbf{P}}\in\mathbb{R}^{D_{p}\times C_{s}}$,
\begin{equation}
\widehat{\mathbf{P}}=\mathbf{P}+ \varphi(\mathbf{L}).
\label{eq:sum2}
\end{equation}
Similarly,
the embedding $\varphi(\mathbf{L})\in\mathbb{R}^{D_{p}\times C_{s}}$ is constructed by a fully-connected layer $\varphi$.
Since $\widehat{\mathbf{P}}$ is context-aware,
we expect that it can enhance the discriminative power of prior boxes to reduce object confusion in few-shot detection.

\textbf{Target OBJ}.
Finally,
we feed $\widehat{\mathbf{P}}$ into target-domain OBJ,
\begin{equation}
\widehat{\mathbf{Y}}=softmax(\widehat{\mathbf{P}}\times\Theta),
\label{eq:targetobj}
\end{equation}
where
$\widehat{\mathbf{Y}}\in\mathbb{R}^{D_{p}\times C_{t}}$ is the target-domain score matrix for classification.
Note that,
target OBJ is shared among different aspect ratios and spatial scales in our design,
with a common parameter matrix $\Theta\in\mathbb{R}^{C_{s}\times C_{t}}$.
One reason is that,
each prior box has combined with its vital contextual fields of different aspect ratios and spatial scales,
i.e.,
each row of $\widehat{\mathbf{P}}$ has become a multi-scale score vector.
Hence,
it is unnecessary to assign an exclusive OBJ on each individual scale.
More importantly,
target domain is few-shot.
The shared OBJ can effectively reduce overfitting in this case.

\textbf{Discussions}.
We further clarify the differences between related works and our Context-Transformer.
\textbf{(1) Few-Shot Learners vs. Context-Transformer}.
Few-shot learners \cite{snell2017prototypical,finn2017model,qi2018low} and our Context-Transformer follow the spirit of learning with little supervision,
in order to effectively generalize model based on few training samples.
However,
most few-shot learners are designed for standard classification tasks.
Hence,
they are often used as a general classifier without taking any detection insights into account.
On the contrary,
our Context-Transformer is a plug-and-play module for object detection.
Via exploiting contexts in a concise transfer framework,
it can adaptively generalize source-domain detector to reduce object confusion in the few-shot target domain.
In fact,
our experiment shows that,
one can fine-tune the pretrained detector with Context-Transformer in the training phase,
and flexibly unload it in the testing phase without much loss of generalization.
All these facts make Context-Transformer a preferable choice for few-shot detection.
\textbf{(2) Non-local Transformer vs. Context-Transformer}.
Non-local Transformer \cite{wang2018non} and our Context-Transformer follow the spirit of attention \cite{vaswani2017attention} for modeling relations.
However,
the following differences make our Context-Transformer a distinct module.
First,
Non-local Transformer does not take any few-shot insights into account.
Hence,
it would not be helpful to reduce training difficulty with little supervision.
Alternatively,
our Context-Transformer leverages source knowledge as guidance to alleviate overfitting in few-shot cases.
Second,
Non-local Transformer is not particularly designed for object detection.
It is simply embedded between two convolution blocks in the standard CNN for space/spacetime modeling.
Alternatively,
our Context-Transformer is developed for few-shot object detection.
We elaborately embed it between source and target OBJs in a SSD-style detection framework,
so that it can tackle object confusions caused by annotation scarcity.
Third,
Non-local Transformer is a self-attention module,
which aims at learning space/spacetime dependencies in general.
Alternatively,
our Context-Transformer is an attention module operated between prior boxes and contextual fields.
It is used to automatically discover important contextual fields for each prior box,
and subsequently aggregate such affinity to enhance OBJ for few-shot detection.
In our experiments,
we compare our Context-Transformer with these related works to show effectiveness and advancement.

\section{Experiments}

To evaluate our approach effectively,
we adapt the popular benchmarks as two challenging settings,
i.e.,
few-shot object detection,
and
incremental few-shot object detection.
More results can be found in our supplementary material.

\subsection{Few-Shot Object Detection}

\textbf{Data Settings}.
First,
we set VOC07+12 as our target-domain task.
The few-shot training set consists of $N$ images (per category) that are randomly sampled from the original train/val set.
Unless stated otherwise,
$N$ is 5 in our experiments.
Second,
we choose a source-domain benchmark for pretraining.
To evaluate the performance of detecting novel categories in the target domain,
we remove 20 categories of COCO that are overlapped with VOC,
and use the rest 60 categories of COCO as source-domain data.
Finally,
we report the results on the official test set of VOC2007,
by mean average precision (mAP) at 0.5 IoU threshold.

\textbf{Implementation Details}.
We choose a recent SSD-style detector \cite{liu2018receptive} as basic architecture,
which is built upon 6 spatial scales (i.e., $38\times38$, $19\times19$, $10\times10$, $5\times5$, $3\times3$, $1\times1$).
For contextual filed construction,
we perform spatial max pooling on the first 4 scales of source-domain score tensors,
where the kernel sizes are 3, 2, 2, 2 and the stride is the same as the kernel size.
The embedding functions in Context-Transformer are residual-style FC layers,
where input and output have the same number of channels.
Finally,
we implement our approach with PyTorch~\cite{paszke2017automatic},
where all the experiments run on 4 TitanXp GPUs.
For pre-training in the source domain,
we follow the details of original SSD-style detectors \cite{liu2018receptive,liu2016ssd}.
For fine-tuning in the target domain,
we set the implementation details
where
the batch size is 64,
the optimization is SGD with momentum 0.9,
the initial learning rate is $4\times 10^{-3}$ (decreased by 10 after 3k and 3.5k iterations),
the weight decay is $5\times 10^{-4}$,
the total number of training iterations is 4k.

\begin{table}[t]
\centering
\resizebox{0.92\columnwidth}{!}{
\begin{tabular}{l|c|c|c|c}
\hline
\hline
Method                 & OBJ (S)             & Context-Transformer                  & OBJ (T)         & mAP \\ \hline
Baseline                 &      $\times$            &          $\times$                &        $\surd$       &   39.4   \\ \hline
\multirow{4}{*}{Ours}    &      $\surd$             &          $\times$                &        $\surd$       &   40.9  \\ %
                         &      $\times$            &          $\surd$                 &        $\surd$       &   41.5 \\ %
                         &      $\surd$             &          $\surd$                 &        $\surd$       &   \textbf{43.8}  \\
                         &      $\surd$             &          $\surd\rightarrow\times$                 &        $\surd$       &   43.4  \\ \hline\hline
\end{tabular}
}
\caption{Source Detection Transfer. Baseline: traditional fine-tuning with target-domain OBJ.
$\surd\rightarrow\times$: We fine-tune the pretrained detector with Context-Transformer in the training phase, and then unload it in the testing phase.}
\label{tab:sdt}
\end{table}

\begin{table}[t]
\centering
\resizebox{.85\columnwidth}{!}{
\begin{tabular}{l|c|l|c}
\hline\hline
Context Construction & mAP & Embedding Layer & mAP \\
\hline
\textit{Without }             &  42.5   & \textit{Without}         &  42.2   \\
\textit{Pool\_avg}            &  43.5   & \textit{FC\_no residual} &  43.0   \\
\textit{Pool\_max }           &  \textbf{43.8}   & \textit{FC\_residual}    &   \textbf{43.8}  \\
\hline\hline
Affinity Discovery   &  mAP   & OBJ (Target)    &  mAP   \\
\hline
\textit{Euclidean}& 43.5 & \textit{Separate }     &  41.4   \\
\textit{Cosine} & \textbf{43.8}   & \textit{Share}          &  \textbf{43.8}   \\
\hline\hline
\end{tabular}
}
\caption{Designs of Context-Transformer.}
\label{tab:sal}
\end{table}

\begin{table}[t]
\centering
\resizebox{.88\columnwidth}{!}{%
\begin{tabular}{l|cccccc}
\hline\hline
No. of Shots $(N)$ & 1 & 2 & 3 & 5 & 10 & all \\ \hline
Baseline        &  21.5       &   27.9     &   33.5      &   39.4      &  49.2    &  80.7    \\ \hline
Ours        & \textbf{27.0}       &   \textbf{30.6}    &   \textbf{36.8}     &   \textbf{43.8}     &   \textbf{51.4}  &   \textbf{81.5}   \\ \hline\hline
\end{tabular}%
}
\caption{Influence of Training Shots.}
\label{tab:ns}

\resizebox{.85\columnwidth}{!}{%
\begin{tabular}{l|ccccc|c}
\hline\hline
Trials No. & 1 & 2 & 3 & 4 & 5  & {mean$\pm$std} \\ \cline{1-6}
Baseline   & 41.7   &  43.1  &  37.9  &  43.5  &  38.2 & (Baseline) \\ \hline
Ours       & \textbf{43.7}   &  \textbf{46.5}  &  \textbf{41.5}  &  \textbf{45.7}  &  \textbf{41.1} & {40.4$\pm$2.0} \\ \hline
Trials No. & 6 & 7 & 8 & 9 & 10 & {mean$\pm$std}  \\ \cline{1-6}
Baseline   & 40.4   &  38.8  & 40.7   & 38.7   &  40.6 &  (Ours) \\ \hline
Ours       & \textbf{44.7}   &  \textbf{41.9}  & \textbf{43.4}   & \textbf{42.4}  &  \textbf{42.3} & {\textbf{43.3$\pm$1.8}} \\ \hline\hline

\end{tabular}%
}
\caption{Influence of Random Trials (5-Shot Case).}
\label{tab:nt}

\resizebox{.88\columnwidth}{!}{%
\begin{tabular}{l|cc|cc}
\hline\hline
\multirow{2}{*}{SSD-Style Framework} & \multicolumn{2}{c|}{\cite{liu2016ssd}}                & \multicolumn{2}{c}{(Liu et al. 2018)}              \\ \cline{2-5}
                          & Baseline & Ours & Baseline & Ours \\ \hline
mAP                       & 35.3 & \textbf{38.7}  &39.4  &    \textbf{43.8}        \\ \hline\hline
\end{tabular}
}
\caption{Influence of SSD-Style Framework.}
\label{tab:backbone}
\end{table}

\textbf{Source Detection Transfer}.
The key design in Source Detection Transfer is OBJ.
To reduce object confusion in the few-shot target domain,
we propose to preserve source OBJ, and embed Context-Transformer between source and target OBJs.
In Table \ref{tab:sdt},
we evaluate the effectiveness of this design,
by comparison with baseline (i.e., traditional fine-tuning with only target OBJ).
First,
our approach outperforms baseline,
by adding target OBJ on top of source OBJ.
It shows that,
preserving source OBJ can alleviate overfitting for few-shot learning.
Second,
our approach outperforms baseline,
by adding target OBJ on top of Context-Transformer.
It shows that,
Context-Transformer can effectively reduce confusion by context learning.
Third,
our approach achieves the best when we embed Context-Transformer between source and target OBJs.
In this case,
Context-Transformer can sufficiently leverage source-domain knowledge to enhance target OBJ.
Note that,
our design only introduces 15.6K extra parameters (i.e., Context-Transformer: 14.4K, target OBJ: 1.2K),
while baseline introduces 2,860K extra parameters by adding target OBJ directly on top of multi-scale convolution features.
It shows that,
our design is of high efficiency.
Finally,
we unload Context-Transformer in the testing,
after applying it in the training.
As expected,
the performance drops marginally,
indicating that Context-Transformer gradually generalizes few-shot detector by learning contexts during training.

\begin{table*}[t]
\centering
\resizebox{2.1\columnwidth}{!}{
 \renewcommand\tabcolsep{1.5pt}
\begin{tabular}{l|cccccccccccccccccccc|c}
\hline\hline
VOC2007 (5-Shot Case) & aero  & bike   & bird  & boat  & bottle & bus   & car  & cat  & chair & cow  & table & dog  & horse & mbike  & person & plant & sheep & sofa  & train & tv   & avg  \\ \hline
Prototype (Snell et al. 2017) & 50.0 & 55.0 & 23.7 & \textbf{26.1} & 8.9    & 54.2 & 71.2 & 41.6 & 29.8  & 23.6 & 34.0  & 30.7 & 46.3  & 53.3  & 60.2   & \textbf{21.9}  & 37.3  & 24.3 & 50.2  & \textbf{54.4} & 39.8 \\
Imprinted (Qi et al. 2018) & 49.1 & 54.8 & 26.0 & 23.5 & 14.7   & 53.0 & 71.2 & 53.0 & 30.4  & 21.0 & 34.0  & 28.6 & 48.1  & \textbf{56.4}  & 63.5   & 21.4  & 39.2  & 32.9 & 45.1  & 50.8 & 40.9  \\
Non-local \cite{wang2018non} &   51.9    &    58.1    &   25.3    &   \textbf{26.1}    &   8.5     &   49.7   &   71.9   &  55.3    &   \textbf{32.3}    &   20.1   &31.9       &   \textbf{32.1}   &   44.7    &    55.8    &    63.7    &   16.2    &  41.7     &   \textbf{33.2}    &   52.4    &   49.9   &    41.0  \\ \hline
Our Context-Transformer   & \textbf{55.4}  & \textbf{59.1}  & \textbf{28.6}  & 23.9  & \textbf{15.9}   & \textbf{58.3}  & \textbf{74.5} & \textbf{57.1} & 31.4  & \textbf{26.0} & \textbf{38.1}  & 31.7 & \textbf{55.8}  & 56.1   & \textbf{64.1}   & 18.1  & \textbf{45.8}  & 33.0  & \textbf{53.2}  & 49.9 & \textbf{43.8} \\ \hline\hline
\end{tabular}
}
\caption{Comparison with Related Learners (Few-Shot Object Detection). We re-implement these learners in our transfer framework,
where we replace our Context-Transformer by them.}
\label{tab:lowshot}
\end{table*}

\textbf{Designs of Context-Transformer}.
We investigate key designs of Context-Transformer in Table \ref{tab:sal}.
(1) Context Construction.
First,
the pooling cases are better.
It shows that,
we do not need to put efforts on every tiny details in the images.
Pooling can effectively reduce the number of contextual fields,
and consequently alleviate learning difficulty in comparison between prior boxes and contexts.
Additionally,
max pooling is slightly better than average pooling.
Hence,
we choose max pooling.
(2) Embedding Functions.
As expected,
Context-Transformer performs better with FC layers,
due to the improvement of learning flexibility.
Additionally,
the residual style is slightly better than the no-residual case,
since it can reduce the risk of random initialization especially for few-shot learning.
Hence,
we choose the residual-style FC layers.
(3) Affinity Discovery.
We compute affinity by two popular similarity metric,
i.e.,
Cosine (dot-product) and Euclidean distance.
The results are comparable,
showing that Context-Transformer is robust to the metric choice.
For simplicity,
we choose Cosine in our paper.
(4) OBJ (Target).
Sharing OBJ (Target) among spatial scales achieves better performance.
This perfectly matches our insight in Section \ref{Context Aggregation},
i.e.,
each prior box has combined with its key contextual fields of various spatial scales,
after learning with Context-Transformer.
Hence,
it is unnecessary to assign an exclusive OBJ (Target) for each scale separately.

\textbf{Influence of Shot and Framework}.
First,
the detection performance tends to be improved as the number of training shots increases in Table \ref{tab:ns}.
Interestingly,
we find that the margin between our approach and baseline tends to decline gradually,
when we have more training shots.
This matches our insight that,
Context-Transformer is preferable to distinguish object confusion caused by low data diversity.
When the number of training samples increases in the target domain,
such few-shot confusion would be alleviated with richer annotations.
But still,
Context-Transformer can model discriminative relations to boost detection in general.
Hence,
our approach also outperforms baseline for all-shot setting.
Second,
our approach exhibits high robustness to random trials (Table \ref{tab:nt}),
where
we run our approach on extra 10 random trials for 5-shot case.
The results show that our approach consistantly outperforms baseline.
Finally,
we build Context-Transformer upon two SSD-style frameworks
\cite{liu2016ssd} and \cite{liu2018receptive}.
In Table \ref{tab:backbone},
our approach significantly outperforms baseline.
The result is better on \cite{liu2018receptive} due to multi-scale dilation.

\textbf{Comparison with Related Learners}.
We compare Context-Transformer with popular few-shot learners \cite{snell2017prototypical,qi2018low} and Non-local Transformer \cite{wang2018non}.
We re-implement these approaches in our transfer framework,
where we replace Context-Transformer with these learners.
More implementation details can be found in our supplementary material.
In Fig. \ref{tab:lowshot},
Context-Transformer outperforms Prototype \cite{snell2017prototypical} and Imprinted \cite{qi2018low},
which are two well-known few-shot classifiers.
It shows that,
general methods may not be sufficient for few-shot detection.
Furthermore,
Context-Transformer outperforms Non-local \cite{wang2018non}.
It shows that,
it is preferable to discover affinity between prior boxes and contextual fields to reduce object confusion,
instead of self-attention among prior boxes.

\begin{table}[t]
\centering
\resizebox{1.0\columnwidth}{!}{
\renewcommand\tabcolsep{3.0pt}
\begin{tabular}{c|c|cc|cc|cc}
\hline\hline
\multicolumn{2}{c|}{\multirow{2}{*}{Before Incremental}}                                         & \multicolumn{2}{c|}{Split1} & \multicolumn{2}{c|}{Split2} & \multicolumn{2}{c}{Split3} \\ \cline{3-8}
\multicolumn{2}{c|}{}                                                                & S      & T     & S      & T     & S      & T     \\ \hline\hline
\multirow{3}{*}{\begin{tabular}[c]{@{}c@{}}S\\ (All)\end{tabular}}     & (Shmelkov et al. 2017)& 67.2         & -            & 68.3         & -            & 69.3         & -            \\
                                                                       & \cite{kang2019few}    & 69.7         & -            & 72.0         & -            & 70.8         & -            \\
                                                                       & Ours  &\textbf{72.9}        & -            & \textbf{73.0}         & -            & \textbf{74.0}         & -            \\ \hline\hline
\multicolumn{2}{c|}{\multirow{2}{*}{After Incremental}}                                          & \multicolumn{2}{c|}{Split1} & \multicolumn{2}{c|}{Split2} & \multicolumn{2}{c}{Split3} \\ \cline{3-8}
\multicolumn{2}{c|}{}                                                                & S      & T     & S      & T    & S      & T     \\ \hline\hline
\multirow{3}{*}{\begin{tabular}[c]{@{}c@{}}S+T\\ (1Shot)\end{tabular}} & (Shmelkov et al. 2017)& 52.5         & 23.9         & 54.0         & 19.2         & 54.6         & 21.4         \\
                                                                       & \cite{kang2019few}   & 66.4         & 14.8         & \textbf{68.2}         & 15.7         & 65.9         & 19.2         \\
                                                                       & Ours  & \textbf{67.4}         & \textbf{34.2}        & 68.1         & \textbf{26.0}         & \textbf{66.8}         & \textbf{29.3}         \\ \hline
\multirow{3}{*}{\begin{tabular}[c]{@{}c@{}}S+T\\ (5Shot)\end{tabular}} & (Shmelkov et al. 2017)& 57.4         & 38.8         & 58.1         & 32.5         & 59.1         & 31.8         \\
                                                                       &  \cite{kang2019few}   & 63.4         & 33.9         & 66.6         & 30.1         & 64.6         & 40.6         \\
                                                                       & Ours  & \textbf{67.3}         & \textbf{44.2}         & \textbf{67.4}         & \textbf{36.3}         & \textbf{67.4}        & \textbf{40.8}         \\ \hline\hline
\end{tabular}
}
\caption{Incremental Few-Shot Object Detection (mAP). S: source-domain classes. T: target-domain classes. More details can be found in Section \ref{Incremental Few-Shot Object Detection}.}
\label{tab:increlowshot}

\end{table}

\begin{figure}[t]
\centering
\includegraphics[width=0.95\columnwidth]{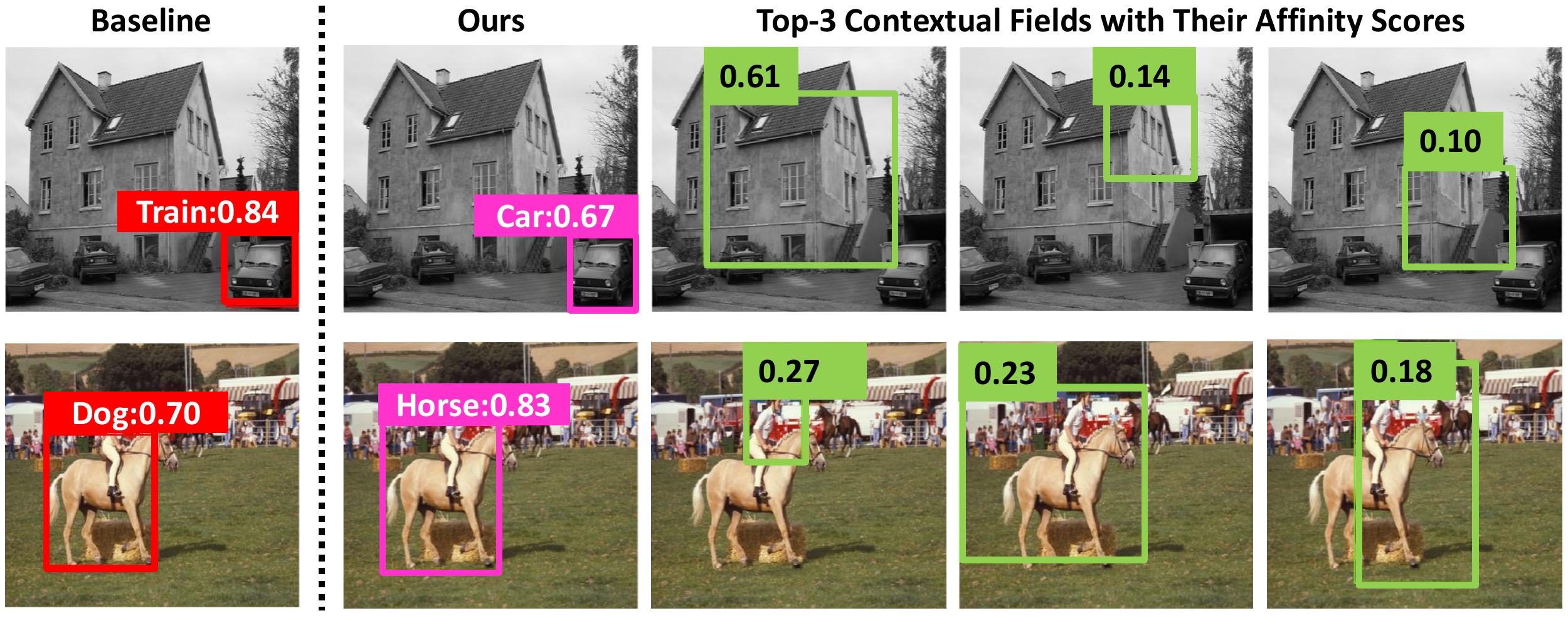}
\caption{Context Affinity.
Context-Transformer can distinguish a \textit{car} from a \textit{train},
when it finds that there is a family house (1st context) with windows (2nd context) and entrance stairs (3rd context).
Similarly,
it can distinguish a \textit{horse} from a \textit{dog},
when it finds that there is a person (1st context) on top of this animal (2nd and 3rd contexts).}
\label{Vis2}

\end{figure}

\begin{figure}[t]
\centering
\includegraphics[width=0.95\columnwidth]{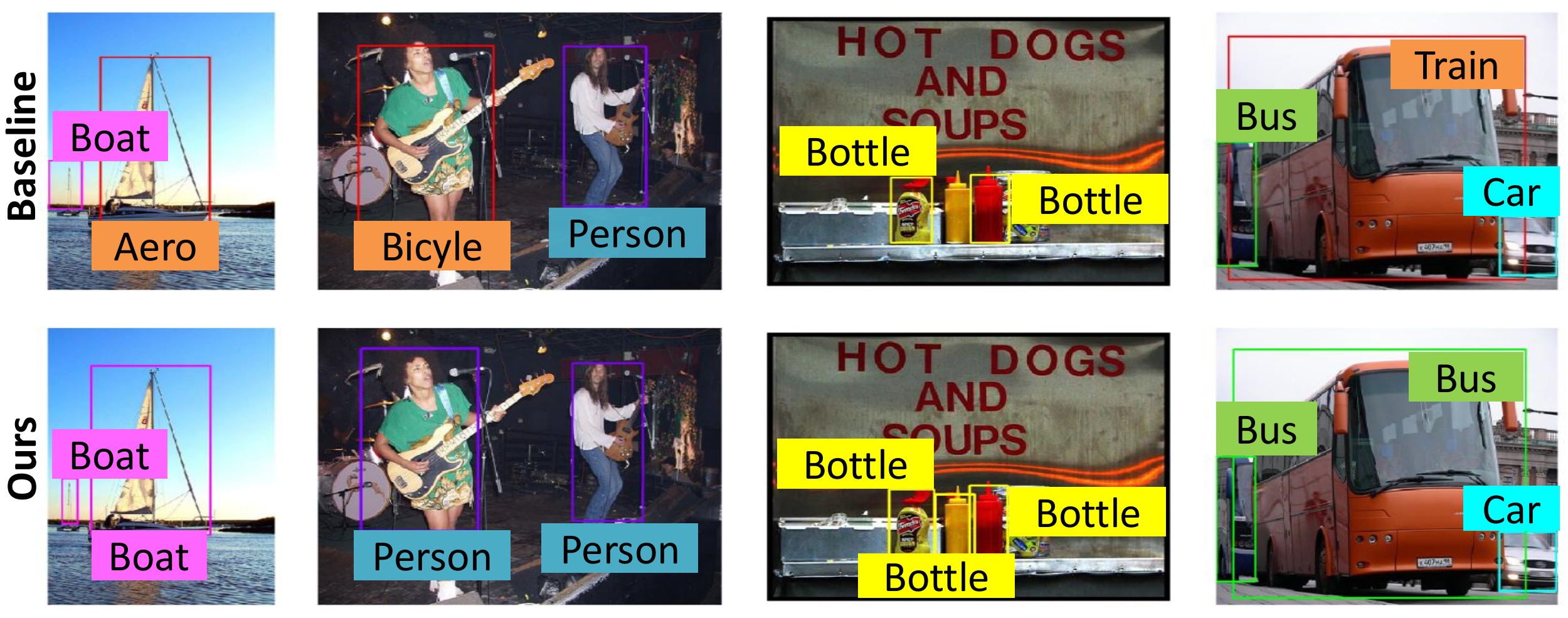}
\caption{Detection Visualization (5-Shot Case).}
\label{Vis3}

\centering
\includegraphics[width=0.95\columnwidth]{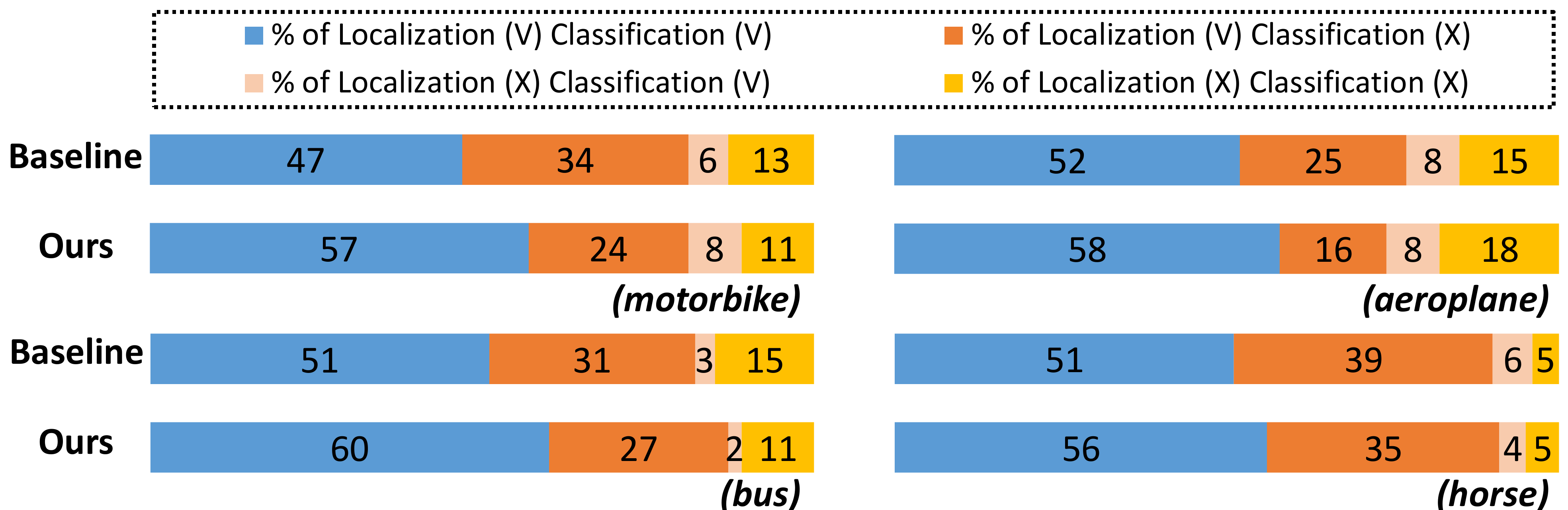}
\caption{Object Confusion (Top-4 Improvement).
Context-Transformer can effectively reduce the correctly-localized but wrongly-classified instances (e.g., baseline vs. ours: 34\% vs. 24\% for \textit{motorbike}),
and promote the correct detection (e.g., baseline vs. ours: 47\% vs. 57\% for \textit{motorbike}).}
\label{Vis1}
\end{figure}

\subsection{Incremental Few-Shot Object Detection}
\label{Incremental Few-Shot Object Detection}

In practice,
many applications refer to an incremental scenario \cite{kang2019few},
i.e.,
the proposed framework should boost few-shot detection in a new target domain,
while maintaining the detection performance in the previous source domain.
Our transfer framework can be straightforwardly extended to address it.
Specifically,
we add a residual-style FC layer on top of the pretrained source-domain OBJ,
which allows us to construct a new source-domain OBJ that can be more compatible with target-domain OBJ.
Then,
we concatenate new source and target OBJs to classify objects in both domains.
More implementation details can be found in our supplementary material.

To evaluate this incremental scenario,
we follow the original data settings of \cite{kang2019few}.
There are 3 data splits that are built upon VOC07+12 (train/val) and VOC07 (test).
For each split,
there are 15 source-domain classes (base) and 5 target-domain classes (novel).
For each target-domain class,
there are only $N$-shot annotated bounding boxes.
We compare our approach with two recent incremental detection approaches,
i.e.,
\cite{shmelkov2017incremental} and \cite{kang2019few}
As shown in Table \ref{tab:increlowshot},
our approach can effectively boost few-shot detection in the target domain as well as maintain the performance in the source domain,
and significantly outperform the state-of-the-art approaches.

\subsection{Visualization}

\textbf{Context Affinity}.
In Fig. \ref{Vis2},
we show top-3 important contextual fields learned by our Context-Transformer.
As we can see,
context affinity can correct object confusion to boost few-shot detection.
Furthermore,
the sum of top-3 affinity scores is over 0.6.
It illustrates that,
Context-Transformer can learn to exploit sparse contextual fields for a prior box,
instead of focusing on every tiny detail in the image.

\textbf{Detection Visualization}.
In Fig. \ref{Vis3},
we show the detection performance for 5-shot case.
One can clearly see that,
our approach correctly detects different objects,
while vanilla fine-tuning introduces large object confusions.

\textbf{Object Confusion Analysis}.
In Fig. \ref{Vis1},
we show top-4 improved categories by our approach.
As expected,
Context-Transformer can largely reduce object confusion and boost performance in few-shot cases.

\section{Conclusion}
In this work,
we propose a Context-Transformer for few-shot object detection.
By attentively exploiting multi-scale contextual fields within a concise transfer framework,
it can effectively distinguish object confusion caused by annotation scarcity.
The extensive results demonstrate the effectiveness of our approach.

\section{Acknowledgements}
This work is partially supported by the National Key Research and Development Program of China (No. 2016YFC1400704), and National Natural Science Foundation of China (61876176, U1713208), Shenzhen Basic Research Program (JCYJ20170818164704758, CXB201104220032A), the Joint Lab of CAS-HK, Shenzhen Institute of Artificial Intelligence and Robotics for Society.

\fontsize{9.0pt}{10.0pt} %
\selectfont
\bibliography{ms}
\bibliographystyle{aaai}
\end{document}


\maketitle

\section{Few-Shot Detection}

\textbf{Contextual Field Construction via Pooling}.
The SSD-style detector is built upon 6 spatial scales (i.e., $38\times38$, $19\times19$, $10\times10$, $5\times5$, $3\times3$, $1\times1$).
Since the spatial resolutions of last two scales are small enough,
it is unnecessary to perform pooling.
In this case,
we perform spatial max pooling on the first 4 scales to construct contextual fields.
As shown in Table \ref{tab:pooling},
pooling is clearly better than no-pooling.
It illustrates that,
pooling achieves high efficiency of learning contextual fields.
Additionally,
the results among different pooling strategies are comparable.
Hence,
we choose the best result (i.e., [3,2,2,2]) in our experiment.

\textbf{Implementation of Related Learners}.
We compare Context-Transformer with well-known few-shot learners \cite{snell2017prototypical,qi2018low} and Non-local Transformer \cite{wang2018non}.
We re-implement these approaches in our transfer framework,
where we replace Context-Transformer with these learners.
(1) Prototype \cite{snell2017prototypical}.
We run prototypical learning on top of source-domain OBJ.
In this case,
target-domain OBJ becomes a metric classifier.
For each category,
its prototype is obtained by averaging over source-domain scores of positive prior boxes (IoU$>$0.5).
Additionally,
we follow \cite{snell2017prototypical} to perform a meta-training procedure (i.e., 20way-5shot) after pretraining source detector,
in order to generalize this classifier in the few-shot target domain.
(2) Imprinted \cite{qi2018low}.
We perform weight imprinting for target-domain OBJ,
based on the score of source-domain OBJ.
For each category,
we normalize source-domain scores of positive prior boxes (IOU$>$0.5),
and average them as the corresponding parameter vector in the target-domain OBJ.
After that,
we fine-tune the whole network in the target domain,
using the same training strategies of our method for fair comparison.
(3) Non-local \cite{wang2018non}.
We embed Non-local Transformer between source and target OBJs,
where affinity is learned among prior boxes without context construction.

\begin{figure*}[t]
\centering
\includegraphics[width=\textwidth]{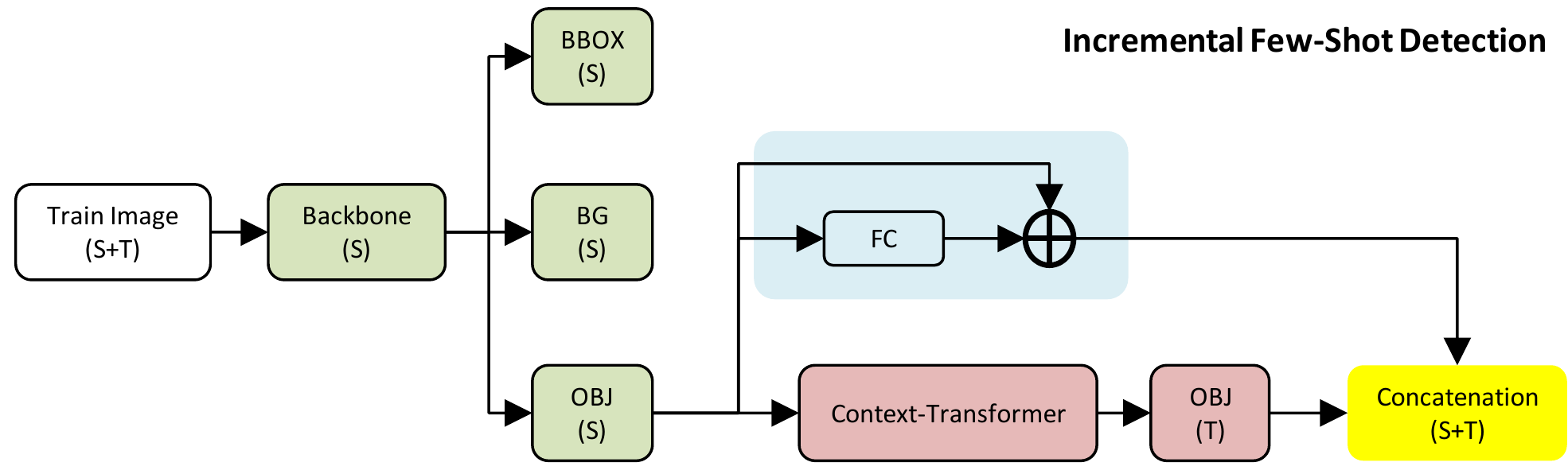}
\caption{Incremental few-shot object detection.
Specifically,
we add a residual-style FC on top of the pretrained source OBJ.
Via such adaptive transformation,
we expect that source OBJ is more compatible with target OBJ.
Finally,
we concatenate the OBJ scores for object classification in both domains.
All other submodules are kept the same as few-shot detection.}
\label{Incre}
\end{figure*}

\begin{table}[t]
\begin{center}
\begin{tabular}{c|cc}
\hline\hline
Pooling & No.of Context Fields & mAP \\
\hline\hline
{[}\textcolor{blue}{2},2,2,2{]} & 2,990 & 43.6  \\
{[}\textcolor{blue}{3},2,2,2{]} & 1,858 & \textbf{43.8}  \\
{[}\textcolor{blue}{5},2,2,2{]} & 1,228 & 43.5  \\
{[}\textcolor{blue}{10},2,2,2{]} & 940 & 43.0  \\
{[}\textcolor{blue}{38},2,2,2{]} & 850 & 43.2  \\
{[}3,\textcolor{blue}{3},2,2{]} & 1,552 & 43.4  \\
{[}3,\textcolor{blue}{5},2,2{]} & 1,354 & 43.1  \\
{[}3,\textcolor{blue}{10},2,2{]} & 1,282 & 43.4  \\
{[}3,\textcolor{blue}{19},2,2{]} & 1,264 & 43.3 \\
{[}3,2,\textcolor{blue}{3},2{]} & 1,804 & 43.2 \\
{[}3,2,\textcolor{blue}{5},2{]} & 1,732 & 43.7 \\
{[}3,2,\textcolor{blue}{10},2{]} & 1,714 & 43.7 \\
{[}3,2,2,\textcolor{blue}{3}{]} & 1,828 & 43.0 \\
{[}3,2,2,\textcolor{blue}{5}{]} & 1,810 & 43.6 \\
{[}-,-,-,-{]} & 11,620 & 42.5 \\
\hline\hline
\end{tabular}
\end{center}
\caption{Contextual Field Construction via Pooling.}
\label{tab:pooling}
\end{table}

\section{Incremental Few-Shot Object Detection}

As shown in Fig. \ref{Incre},
we extend our approach with the following designs to tackle the incremental few-shot scenario.

\textbf{Architecture Adaptation}.
Since Backbone, BBOX and BG are shared between source and target domains,
these submodules are kept the same as before.
Alternatively,
we add a residual-style FC on top of the pretrained source OBJ,
which leads to a new source OBJ.
Via such adaptive transformation,
we expect that the new source OBJ is more compatible with target OBJ.
Finally,
we concatenate the prediction score together to classify objects in both domains.

\textbf{Training Practices}.
First,
we pretrain the source detector as before.
Second,
we sample training images from source-domain benchmark.
For each category,
the number of training shots is the same as the one in the target domain,
in order to alleviate data bias.
Subsequently,
we group this few-shot source data and the available few-shot target data as a training set,
and fine-tune the architecture in Fig. \ref{Incre}.
As a result,
we can generalize the detection performance in the target domain as well as reduce forgetting in the source domain.

\textbf{Implementation Details}.
Following the data setting in \cite{kang2019few},
we perform model pretraining on VOC07+12 training set (only 15 source-domain classes defined by \cite{kang2019few}).
The pretraining details is the same as the original SSD-style detectors \cite{liu2016ssd,liu2018receptive}.
For fine-tuning in the target domain,
we use the implementation details as follows.
For $N$-shot setting,
there are only $N$ annotated bounding boxes available for each target-domain category.
The batch size is 64 for 5 shot case (or 20 for 1 shot case).
The optimization is SGD with momentum 0.9.
The initial learning rate is $4\times10^{-3}$ (decreased by 10 after 0.3k/0.6k iterations for 1 shot/5 shot case).
The weight decay is $5\times10^{-4}$.
The total number of training iterations is 1k for 5 shot case (or 0.5k for 1 shot case).
Additionally,
under the same data setting,
we use the published code to re-implement \cite{shmelkov2017incremental},
and pick the result of \cite{kang2019few} from its original paper,
for the state-of-the-art comparison.

\begin{figure*}[t]
\centering
\subfigure[Source-Domain Object Knowledge (No Fine-Tuning)]{\includegraphics[width=1\textwidth,trim=0 0 0 28,clip]{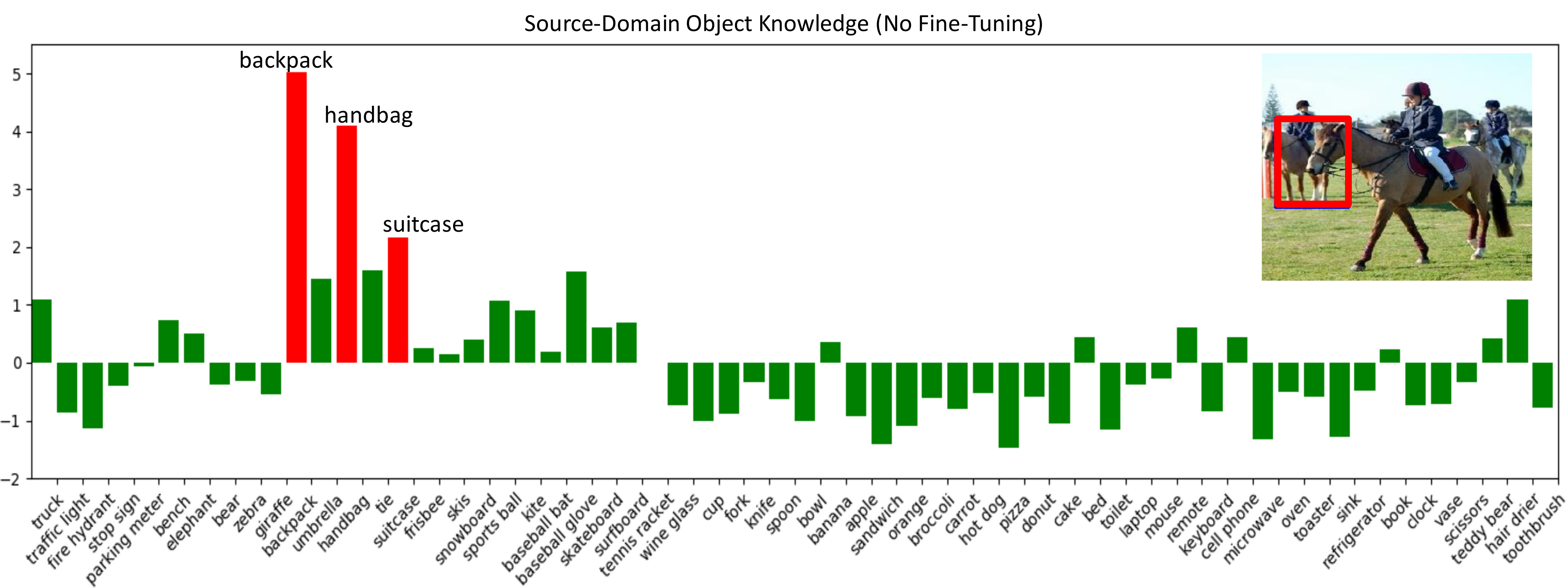}}
\subfigure[Source-Domain Object Knowledge (Fine-Tuning without Context-Transformer)]{\includegraphics[width=1\textwidth,trim=0 0 0 28,clip]{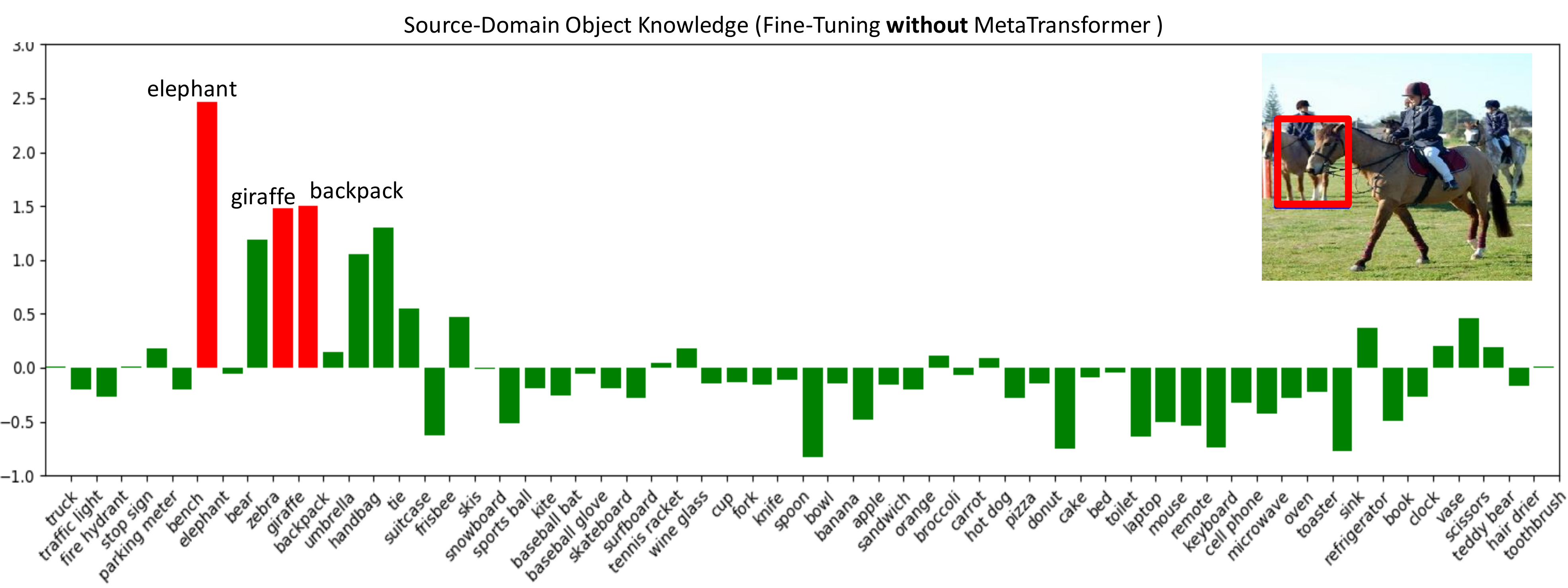}}
\subfigure[Source-Domain Object Knowledge (Fine-Tuning with Context-Transformer)]{\includegraphics[width=1\textwidth,trim=0 0 0 28,clip]{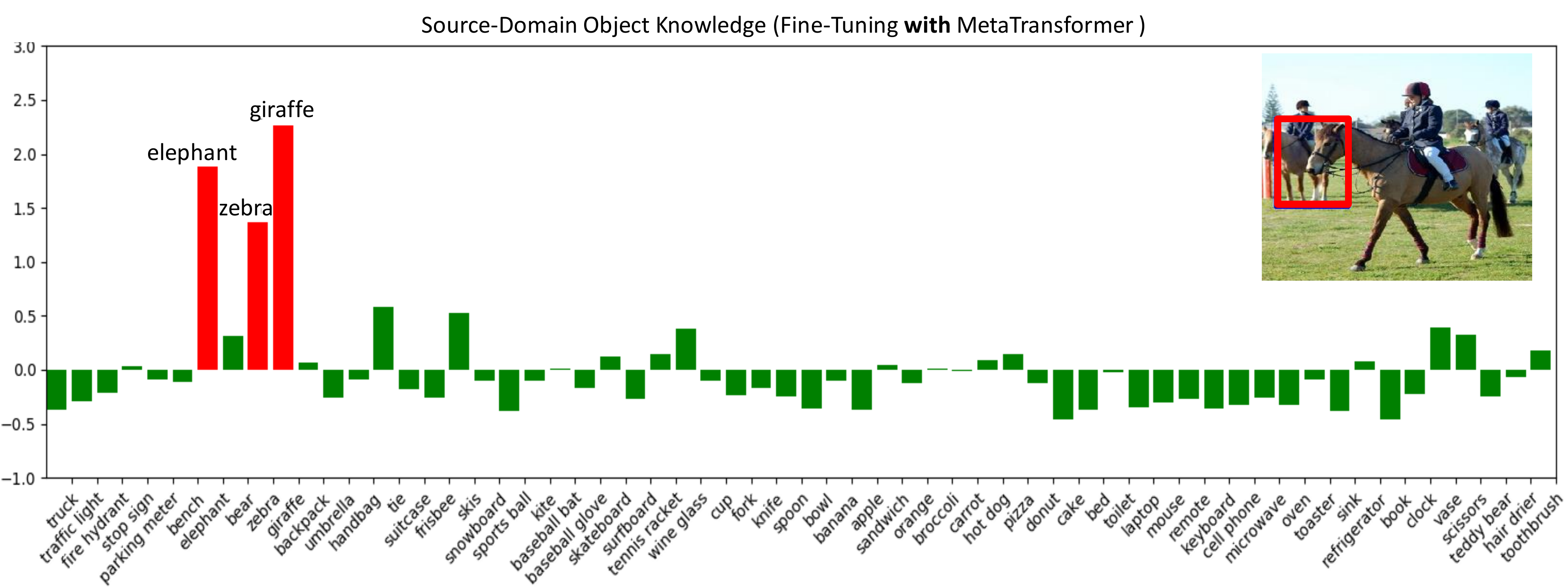}}
\caption{Source-domain object knowledge for a prior box in the target domain image.
For a prior box (red box),
we extract the prediction score vector from source-domain OBJ,
and highlight its top-3 activated source object classes (red bars).
We can clearly see that,
source-domain OBJ produces more discriminative semantic clues after fine-tuning with our Context-Transformer,
e.g.,
\textit{horse} is more similar to \textit{giraffe}, \textit{zebra} and \textit{elephant} in terms of appearance.
It illustrates that,
Context-Transformer can effectively generalize source-domain detector for few-shot detection in the target domain.}
\vspace{-1cm}
\label{skShow}
\end{figure*}

\begin{figure*}[t]
\centering
\vspace{-0.3cm}
\includegraphics[width=0.9\textwidth]{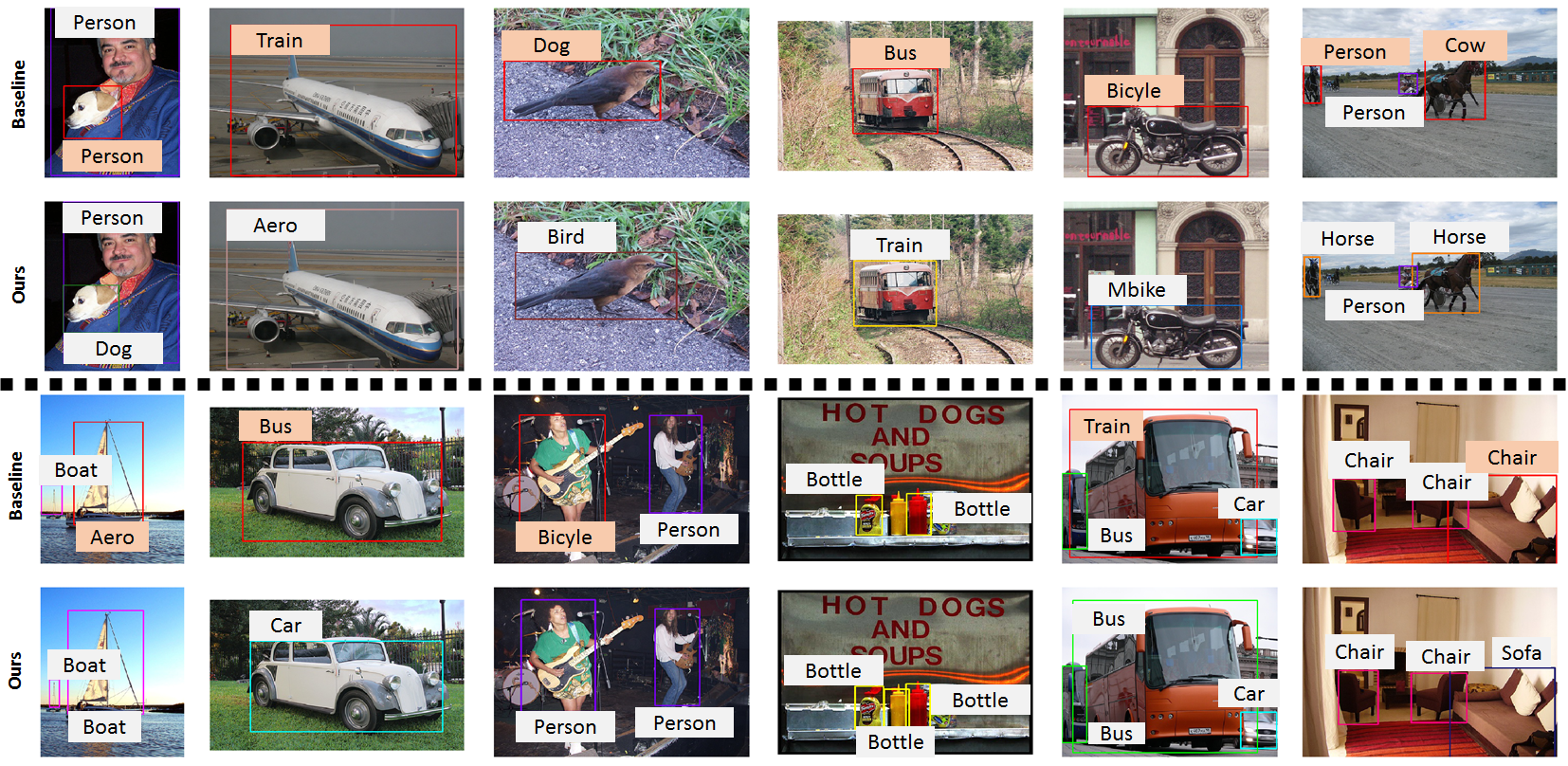}
\vspace{-0.3cm}
\caption{Detection Visualization (5-Shot Case).}
\label{VisShow}
\vspace{0.5cm}
\centering
\includegraphics[width=0.9\textwidth]{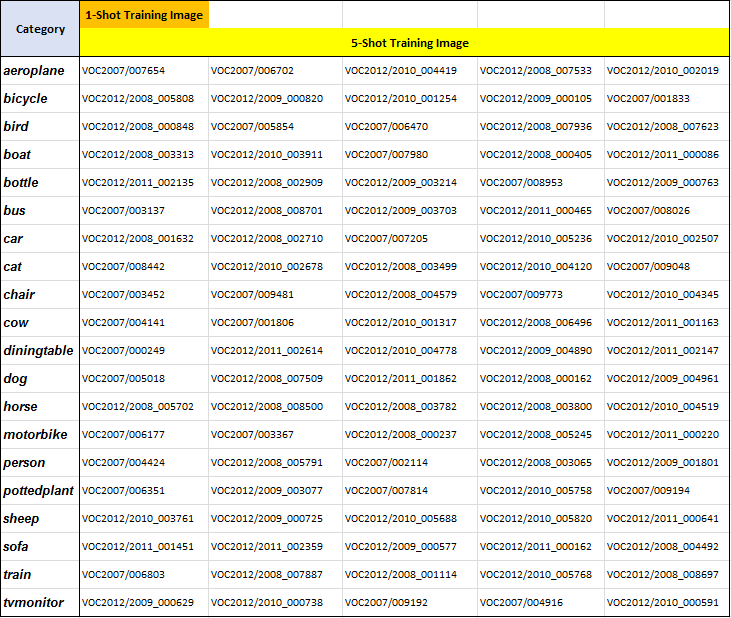}
\vspace{-0.3cm}
\caption{Index of training samples in our experiments for few-shot object detection.}
\vspace{-1cm}
\label{indexShow}
\end{figure*}

\section{Visualization}

\textbf{Source-Domain Object Knowledge}.
Since we use source-domain score as guidance to perform Context-Transformer,
we further investigate this source-domain object knowledge.
Specifically,
we extract the prediction score vector from source-domain OBJ,
w.r.t.,
a prior box in the target-domain image (e.g., the red box in Fig. \ref{skShow}).
We make comparisons among 3 cases,
i.e.,
original score (without fine-tuning),
adjusted score (fine-tuning without Context-Transformer),
adjusted score (fine-tuning with Context-Transformer).
We can see in Fig. \ref{skShow} that,
top-3 activated source objects of original score vector are \textit{backpack}, \textit{handbag}, \textit{suitcase}.
Apparently,
these items are not discriminative enough to recognize \textit{horse},
even though they have similar color and shape to \textit{horse} in that prior box.
Via fine-tuning without Context-Transformer,
source OBJ attempts to adjust the prediction score,
which activates \textit{elephant} and \textit{giraffe} in the source-domain classes.
However,
it is still not sufficient to recognize \textit{horse}.
In fact,
this approach mistakenly classifies \textit{horse} as \textit{cow}.
Finally,
fine-tuning with our Context-Transformer produces the most relevant clues to recognize \textit{horse},
i.e.,
\textit{giraffe}, \textit{zebra} and \textit{elephant}.
It illustrates that,
Context-Transformer can effectively generalize source-domain detector for few-shot detection in the target domain.

\textbf{Detection Visualization}.
In Fig. \ref{VisShow},
we further show the detection performance of 5-shot case.
We can clearly see that,
our approach correctly detects different objects,
while baseline introduces large object confusions.

\textbf{Training Samples}.
To evaluate if our approach can detect novel categories in the target domain,
we remove 20 categories of COCO that are overlapped with VOC, and use the rest 60 categories of COCO as source-domain dataset.
Specifically,
we remove the annotations in an image that belong to the overlapped categories and completely drop such image if there is no annotation left.
After that,
source-domain dataset consists of 98,446 images in the train/val set and 4,166 images in the test set.
We use the train/val set for model pretraining.
Furthermore,
we provide readers with the selected training images of 1-shot and 5-shot settings in the target-domain VOC dataset (Fig. \ref{indexShow}),
for better reproducing our results.

{\small
\bibliographystyle{aaai}
\bibliography{supplement}
}